\documentclass[runningheads]{llncs}
\usepackage[T1]{fontenc}
\usepackage{graphicx}
\usepackage{float} 
\usepackage{hyperref} 
\usepackage{booktabs} 
\usepackage{tikz}
\usetikzlibrary{tikzmark}

\usepackage{amsmath}
\usepackage{amsfonts}
\usepackage{amssymb}
\usepackage{mathtools}
\usepackage{xcolor}

\begin{document}
\title{Investigating the Interplay of Parameterization and Optimizer in Gradient-Free Topology Optimization: A Cantilever Beam Case Study}
\titlerunning{Investigating the Interplay of Parameterization and Optimizer in TO}
%
%
\author{Jelle Westra \and
Iv\'{a}n Olarte Rodr\'{i}guez\orcidID{0009-0005-0748-9069} \and
Niki van Stein\orcidID{0000-0002-0013-7969} \and
Thomas Bäck\orcidID{0000-0001-6768-1478} \and
Elena Raponi\orcidID{0000-0001-6841-7409}}
%
\authorrunning{J. Westra et al.}
%
\institute{Leiden Institute of Advanced Computer Science, Leiden University, Einsteinweg 55
2333 CC Leiden, The Netherlands \\
\email{\{i.olarte.rodriguez, n.van.stein, t.h.w.baeck, e.raponi\}@liacs.leidenuniv.nl}}
\maketitle              
\begin{abstract}

Gradient-free black-box optimization (BBO) is widely used in engineering design and provides a flexible framework for topology optimization (TO), enabling the discovery of high-performing structural designs without requiring gradient information from simulations. Yet, its success depends on two key choices: the geometric parameterization defining the search space and the optimizer exploring it. 

This study investigates this interplay through a compliance minimization problem for a cantilever beam subject to a connectivity constraint. We benchmark three geometric parameterizations, each combined with three representative BBO algorithms: differential evolution, covariance matrix adaptation evolution strategy, and heteroscedastic evolutionary Bayesian optimization, across 10D, 20D, and 50D design spaces.


Results reveal that parameterization quality has a stronger influence on optimization performance than optimizer choice: a well-structured parameterization enables robust and competitive performance across algorithms, whereas weaker representations increase optimizer dependency. 
Overall, this study highlights the dominant role of geometric parameterization in practical BBO-based TO and shows that algorithm performance and selection cannot be fairly assessed without accounting for the induced design space.

\keywords{Gradient-free optimization \and Topology optimization \and Geometric parametrization \and Constrained optimization \and Bayesian optimization.}
\end{abstract}


\section{Introduction}
The performance of heuristic optimization algorithms is shaped not only by their design but also by how the optimization problem itself is modeled and parameterized \cite{back1997handbook}. Despite this, the interplay between problem formulation and solver behavior remains comparatively underexplored. In practice, much of the effort in heuristic optimization focuses on improving algorithm design, selection, and configuration \cite{kerschkeAutomatedAlgorithmSelection2019a,hutterAutomatedMachineLearning2019a}, while the impact of parameterization, constraint modeling, and geometric representation on solver performance receives comparatively less attention. 

The interplay between problem formulation and solver behavior is particularly evident in engineering design, and particularly in the field of Topology Optimization (TO), which concerns finding the optimal material distribution within a bounded domain for a given objective \cite{bendsoe2013topology}.
While gradient-based methods such as Solid Isotropic Material with Penalization method (SIMP) \cite{bendsoe1989optimal}, combined with Method of Moving Asymptotes (MMA) \cite{svanberg1987method}, have dominated the structural TO field due to their efficiency and maturity, they rely on differentiable and well-behaved objectives. In contrast, many practical TO problems, for example those involving discontinuities, non-smooth physical responses returned by numerical finite element (FE) simulations, require heuristic or gradient-free black-box optimization (BBO) methods \cite{bujny2018identification}. However, because numerical FE solvers are computationally expensive, BBO methods used in such contexts must operate under tight evaluation budgets. This naturally favors surrogate-based approaches, such as Bayesian Optimization (BO) \cite{garnett2023bayesian} and Surrogate-Assisted Evolutionary Algorithms (SAEAs) \cite{jinSurrogateassistedEvolutionaryComputation2011}, which approximate the expensive simulation response to guide the search more efficiently.
Unlike gradient-based methods that can handle design spaces with thousands of variables (as in SIMP), surrogate-based BBO methods typically struggle with high-dimensional search spaces \cite{forresterEngineeringDesignSurrogate2008}. Building reliable surrogate models requires a number of samples that grows exponentially with dimensionality, making high-dimensional modeling infeasible under limited budgets. Therefore, in simulation-driven TO, the tractability of surrogate-based optimization hinges on reducing the effective dimensionality of the search space through carefully crafted geometry parameterizations.

An effective reduction in search space dimensionality through carefully designed parametrization comes at a cost: the chosen parameterization restricts the freedom of design and fundamentally shapes the fitness landscape \cite{mdobook}. For example, a parameterization based on beams will inherently produce beam-like structures, while one based on hexagonal tiling is restricted to tiled grids. Consequently, even when parameterizations share the same dimensionality, they induce different fitness landscapes. This directly impacts the performance of different BBO algorithms. 
While such methods can handle nonconvex, noisy, and discontinuous objectives, their sensitivity to landscape characteristics makes their performance highly dependent on the chosen parameterization.


In this paper, we investigate the interplay between problem parametrization and algorithm selection through a structural topology optimization case study: minimizing compliance in a horizontal cantilever beam under volume and connectivity constraints. Although this setup admits a closed-form formulation, we treat it as an abstraction of high-fidelity optimization scenarios, where objective evaluations rely on costly numerical FE simulations, so that the discussion generalizes to such scenarios. We made this choice because finite element simulations often depend on licensed software and expert knowledge, while our goal is to ensure reproducibility and accessibility.
We systematically compare three parameterizations of varying expressiveness: straight beams, curved beams, and a hexagonal tiling, against three representative BBO algorithms: Differential Evolution (DE) \cite{storn1997differential}, Covariance Matrix Adaptation Evolution Strategy (CMA-ES) \cite{hansen2001completely}, and Heteroscedastic Evolutionary Bayesian Optimization (HEBO) \cite{cowen2022hebo}, which were chosen to represent a diverse range of optimization strategies: DE is a local search method driven by variation and selection within a population; CMA-ES adapts a probabilistic model of the population distribution; and HEBO builds a global surrogate model of the objective landscape.


Our results reveal that parameterization quality has a stronger influence on optimization performance than the choice of optimizer itself. Well-designed parameterizations consistently yield competitive solutions across all tested optimizers, while poorly structured ones create challenging, multi-modal landscapes where performance depends heavily on the solver's search dynamics. This suggests that the design of the search space, through parameterization, is at least as critical as algorithm selection in achieving robust and high-quality results.

\section{Related Work}

\subsection{Black-Box Optimization}
Black-box optimization refers to the class of methods that aim to optimize an objective function accessible only through evaluations, without analytical gradients or structural information. Over the past decades, a wide range of BBO approaches has been developed, from early stochastic search heuristics to more advanced metaheuristics and surrogate-based algorithms. Evolutionary algorithms (EAs) \cite{back1996evolutionary} form one of the most established families, relying on populations of candidate solutions and biologically inspired operators such as mutation, crossover, and selection to balance exploration and exploitation. Classical examples include DE \cite{storn1997differential}, which perturbs individuals through differential mutation, and CMA-ES \cite{hansen2001completely}, which adapts the covariance structure of a Gaussian search distribution to the underlying landscape. These methods are known for their robustness on noisy or rugged fitness landscapes, though their population-based search can require many evaluations. 

In contrast, BO \cite{garnett2023bayesian} uses probabilistic surrogate models---typically Gaussian Processes for continuous search spaces---to guide the search toward promising regions via acquisition functions. Recent extensions such as HEBO \cite{cowen2022hebo} have made BO more scalable to high-dimensional and complex search spaces, often yielding superior performance under stringent evaluation budgets. Together, these families exemplify complementary philosophies in BBO: EAs emphasize adaptive sampling through populations, while BO leverages statistical modeling to prioritize informative evaluations. Accordingly, we select these three optimizers as representative BBO approaches to investigate how algorithm performance interacts with problem parametrization in the context of TO.

\subsection{Geometry Parametrization in Topology Optimization}
TO aims to automatically determine the optimal material layout within a prescribed design domain to achieve the best structural performance. In the cantilever beam test case analyzed in this work, it aims at minimizing compliance (i.e., maximizing stiffness) under given constraints. Unlike traditional shape or size optimization, TO allows for the creation of entirely new structural topologies, making it a powerful design tool in engineering. However, its effectiveness strongly depends on how the geometry is parameterized, as this determines both the dimensionality and the characteristics of the resulting optimization problem.

Early methods such as SIMP \cite{bendsoe1989optimal} represent the design domain as a discretized FE grid, where each cell holds a continuous material density parameter. The optimization then proceeds by iteratively updating these parameters, effectively searching over a high-dimensional space of material distributions. However, applying EAs on such high-dimensional space requires tens of thousands of evaluations \cite{chapman1994genetic}. Opposed to rectangular grids, a Honeycomb Tiling (HT) parameterization has been shown overcome checker-boarding patterns and stiffness singularity regions \cite{saxena2003honeycomb}. More recently, explicit geometry parameterizations have gained traction for reducing design dimensionality. The Moving Morphable Components (MMC) approach \cite{guo2014doing} represents structures using a set of parameterized beam-like components, allowing for efficient representation of beam-like structure with fewer variables. These parameterizations have been successfully integrated for vehicle crashworthiness with various BBO methods, like EAs \cite{bujny2018identification} and BO \cite{raponi2019kriging}, demonstrating their effectiveness in operating within problems where gradient-based methods are not reliable. Curved MMCs \cite{guo2016explicit} extends MMCs by allowing for more complex geometries through additional deformation parameters. Attempts have been made to capture problem characteristics imposed by these different parameterizations through Exploratory Landscape Analysis \cite{mersmann2011exploratory}, but a comprehensive understanding of how these features interact with different optimizers remains elusive. Mostly, these studies have focused on continuous optimization benchmark problems, such as the Black-Box Optimization Benchmarking (BBOB) test suite from the COmparing Continuous Optimizers (COCO) platform \cite{hansen2009real}, yet attempts that extend these to complex real-world problems are rare \cite{longGeneratingCheapRepresentative2024}.

\section{Problem Formulation}

\subsection{Parametrization}

\begin{figure}[t]
    \centering
    \includegraphics[width=\linewidth]{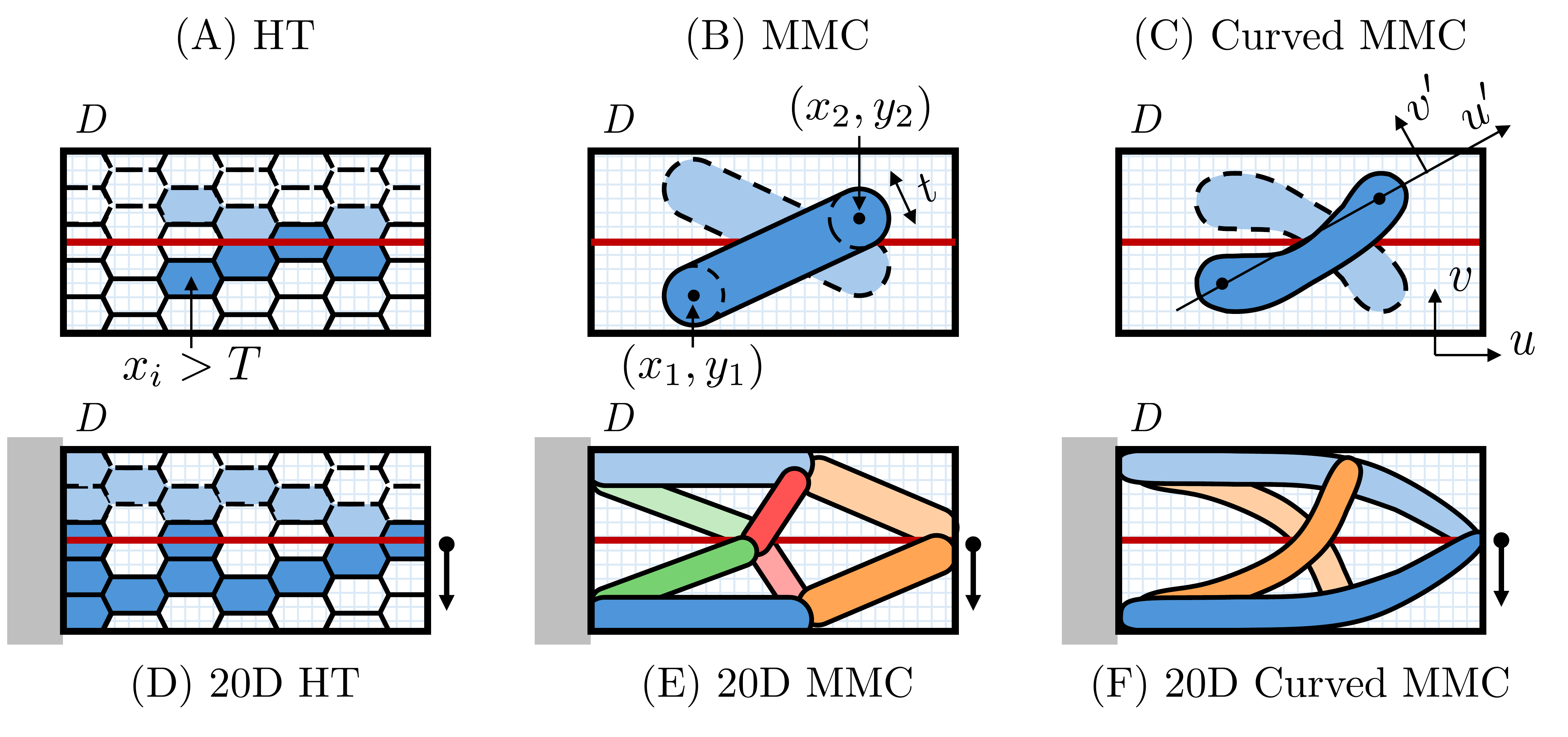}
    \caption{
    Overview of the parameterizations considered in this work: (A) HT, (B) MMCs, and (C) Curved MMCs. Dark-blue regions indicate material directly generated by the parameterization, while light-blue regions show the mirrored material domain enforced by symmetry about the red midline. (D–F) illustrate examples of 20D parameterizations for the full cantilever: (D) HT, (E) MMCs, and (F) Curved MMCs.
    }
    \label{fig:parameterizations}
\end{figure}

Three different geometrical parameterizations are considered in this work: HT, MMCs, and Curved MMCs (\autoref{fig:parameterizations}).  We define a $d$-dimensional parameterization as $\mathcal{P}:[0,1]^d\to\Omega$, where $\Omega\subseteq D$ is the material domain within the design domain $D$, and $[0,1]^d$ is the unit parameter space. Since we only consider a bounded two-dimensional design domain $D\subset\mathbb{R}^2$, voids (shaded white) represent lack of material and is the complement of the material domain: $D\backslash \Omega$. 
For all generated geometries, the design is mirrored with respect to the horizontal midline, as illustrated in \autoref{fig:parameterizations}.

\subsubsection{Honeycomb Tiling}
HT (\autoref{fig:parameterizations}A) consists of a hexagonal grid of tiles that are activated if its corresponding design variable exceeds a threshold $T$ ($T = 1/2$ in this study): 
\begin{equation}
    x_i > T \implies \Omega_i\subseteq \mathcal{P}(\mathbf{x})
\end{equation}
where $x_i$ and $\Omega_i$ denote the $i$-th design variable and its associated material domain respectively. This means the HT has one Degree of Freedom (DoF) per tile.

\subsubsection{MMC}
Unlike previous applications of the MMC beams \cite{raponi2019kriging}, we choose to parameterize the beam using endpoints (\autoref{fig:parameterizations}B), opposed to an angular parameterization. In both ways, a quintuplet of parameters is used to describe the geometry of the beam. For an \textit{endpoints parameterization} the parameters are given by the two endpoints positions $(x_1,y_1)$ and $(x_2,y_2)$, and a thickness $t$.
Given that we need five parameters to describe the geometry of a beam in $\mathbb{R}^2$, the MMC parameterization has 5 DoF per beam. We explicitly choose this endpoints representation to prevent geometry being defined outside of the design domain.
Also, contrary to the original formulation using a implicit hyperellipse function, we generate capsule shape formed by connecting to semi-circles with diameter $t$, with the origin of the semi-circle being defined by the endpoint (see \autoref{fig:parameterizations}B). 

\subsubsection{Curved MMC}
In the Curved MMC parameterization, five additional deformation parameters are added to the (regular, straight-beam) MMCs. Following the definition by Guo et al. \cite{guo2016explicit}, two additional thickness parameters are included, together with three additional beam centerline deformation parameters. We define the deformation of material points in the local coordinate frame $(u',v')$ aligned with the beam orientation as:
\begin{equation}
    \begin{pmatrix}u' \\ v'\end{pmatrix}
    \mapsto
    \begin{pmatrix}u' \\ \phi(u')v' + v'_\text{CL}(u')\end{pmatrix}.
\end{equation}
This represents an affine transformation of $v'$ scaled by $\phi(u')$ and shifted by $v'_\text{CL}(u')$, both as function of $u'$.

A total of three thickness parameters $(t_L, t_M, t_R)$ define the beam thickness at the left, middle (original), and right points of the beam, respectively. The intermediate thickness of the beam is described by the parabolic interpolation of these three thicknesses:
\begin{equation}
    \phi(u') = (t_L + t_R - 2t_M)/4 \cdot (2u'/l)^2 + (t_R - t_L)/4 \cdot (2u'/l) + t_M
\end{equation}
The centerline is deformed using a sinusoidal function. Since we make use of an endpoints representation of the beam, we make a slight adjustment to the original formulation \cite{guo2016explicit}. Originally, length $l$ is decomposed into a left- and right-length, analogous to the thickness. This is de facto the same as adding a phase to the sinusoidal deformation; both effectively shift the center point among the centerline of the beam. Hence we swap this length parameter decomposition for a phase variable in the sine deformation $c$. The full deformation of the centerline can be defined as:
\begin{equation}
    v'_\text{CL}(u') =  a \sin(b(u' + c))
\end{equation}
Additional to phase-shift $c$, the amplitude $a$ and frequency $b$ of the sine deformation are the other two deformation parameters. In total, the Curved MMC parameterization has 10 DoF per beam, double that of the regular MMCs.

\subsection{Optimization Problem}
\begin{figure}[h]
    \centering
    \includegraphics[width=\linewidth]{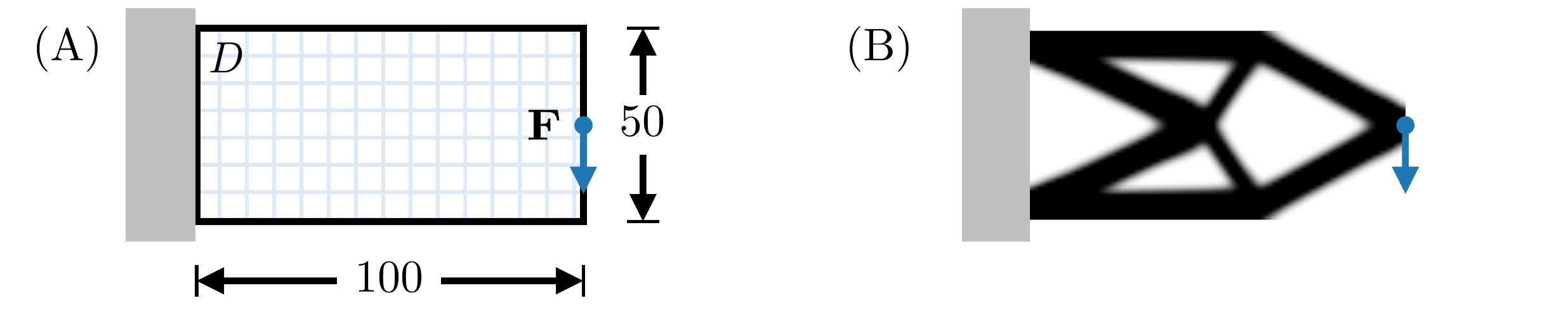}
    \caption{(A) A schematic overview of the studied problem, showing the design domain $D$ and application of load $\mathbf{F}$. (B) The optimal topology obtained by SIMP for the same mesh using \cite{andreassen2011efficient} (adapted from \cite{bujny2020level}).}
    \label{fig:problem-definition}
\end{figure}

In line with previous work \cite{bujny2018identification,raponi2019kriging}, we study a horizontal cantilever beam problem. The material domain is rectangular with a 2:1 ratio, its left-hand side is fixed, and a load is applied to the center of the right-hand side (\autoref{fig:problem-definition}). This domain is subdivided into $100\times50$ square elements to form the mesh of the FE Method (FEM) setup. The goal is to minimize the compliance of the binary material distribution with a 50\% volume constraint. To compute this compliance, the structure is analyzed using a plane-stress, linear elastic material model, in accordance with \cite{serhat2019lamination}.
Additional to the volume constraint, to limit unnecessary calls to the FEM simulation, a connectivity constraint is introduced which is violated if the design is disconnected. This leads to the general problem definition:
\begin{equation}
\label{eq:opt-problem-def}
\begin{cases}
\text{minimize} &f(\mathbf{x}) \\
\text{subject to} & g_1(\mathbf{x}) = \mathcal{V}(\mathbf{x}) - V_\text{max} \leq 0 \\
            & g_2(\mathbf{x}) = \mathcal{C}(\mathbf{x}) \leq 0
\end{cases}
\end{equation}
where $f$ denotes the structural compliance, $\mathcal{V}$ the volume of the material domain, $V_{\text{max}}$ (we set $V_{\text{max}} = 50\%$ of $D$), the volume threshold, and $\mathcal{C}(\mathbf{x})$ is a connectivity constraint, further described below. Herein, the compliance $f$ and constraints $g$ are implicit functions of design vector $\mathbf{x}$; according to the material domain generated by the parameterization.

We propose a parameterization-independent connectivity constraint $\mathcal{C}(\mathbf{x})$ based on the shortest distance required to achieve full structural connectivity. In this context, fully connected means that any point within the design domain is reachable from any other point, and that the structure maintains a continuous connection to the prescribed boundary conditions. Violations of this connectivity can lead to numerical instabilities in the FE solver and may produce spurious or uninformative objective responses.

\begin{figure}[t]
    \centering
    \includegraphics[width=0.7\linewidth]{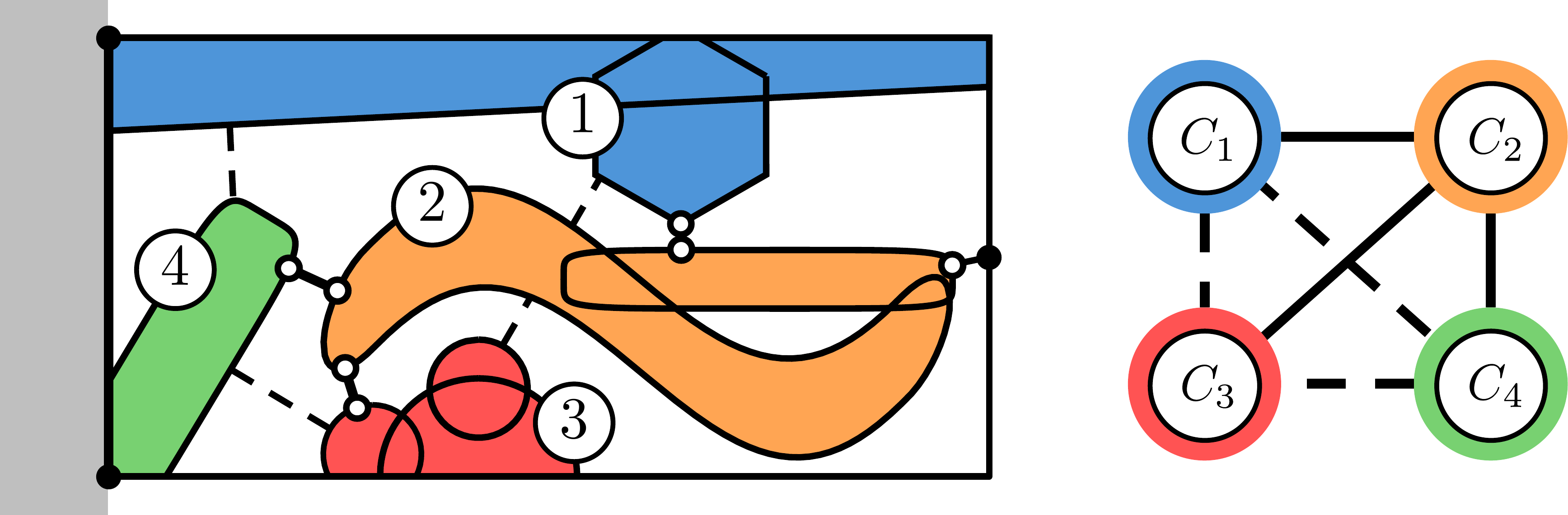}
    \caption{An example for calculating the least distance to connect four connected components of geometry to each other and to edge- and load-point boundary geometry.}
    \label{fig:connectivity}
\end{figure}

Let $C_i$ denote the connected components (shown in different colors in \autoref{fig:connectivity}), as a partition of the material domain $\Omega$:
\begin{equation}
    C_i \cap C_j = \emptyset \text{ if }i\neq j 
    \quad\text{and}\quad
    \Omega = \bigcup C_i.
\end{equation}
We define the shortest Euclidean distance to connect a pair of geometric components as:
\begin{equation}
        d_\text{min}(C_i, C_j) = \inf \{
    \|\mathbf{x}_i - \mathbf{x}_j\|_2 :
    \mathbf{x}_i\in C_i \text{ and }
    \mathbf{x}_j\in C_j
\}.
\end{equation}
Hence, the cost matrix of connecting islands (connected components) pair-wise can be obtained this way:
\begin{equation}
    (\mathbf{W}_C )_{ij} =
    (\mathbf{W}_C )_{ji} = 
    d_\text{min} (C_i,C_j).
\end{equation}
The most economical way of connecting the structure can be found by finding the Minimum Spanning Tree (MST) of this graph representation \cite{kruskal1956shortest}. The MST transforms the weight matrix $\mathbf{W}_C$ into a new matrix $\mathbf{W}_{C,\text{MST}}$, which retains only the minimal-cost set of connections required to keep the graph fully connected. The discarded connections are shown as dashed lines in \autoref{fig:connectivity}.

Hence, we define our connectivity constraint as:
\begin{equation}
    \mathcal{C}(\mathbf{x}) = 
\underbrace{
    \sum_{\Gamma_k \in \Omega_B}
    \min_j d_\text{min} (\Gamma_k, C_j)
}_\text{components to boundaries} + 
\underbrace{
    \sum_{(i,j)}(\mathbf{W}_{C,\text{MST}}(\mathbf{x}))_{ij}
}_{\text{inter-component}},
\end{equation}
where $\Omega_B =\{\Gamma_k\}$ denotes the set of boundary geometries such as the loading point and the left-side support in our problem. A minimum distance constraint $d_\text{min} \geq 0$ implies a lower bound $\mathcal{C} \geq 0$.

To bring both the volume and the connectivity penalization on the same order of magnitude, we define normalized penalization as follows:
\begin{equation}
\tilde{g}_1(\mathbf{x}) = \frac{1}{V_\text{domain}} \max\{
    (\mathcal{V}(\mathbf{x}) - V_\text{max}), 0
\}
\end{equation}
where $V_\text{domain}$ denotes the total volume of the domain. We also define
\begin{equation}
\tilde{g}_2(\mathbf{x}) = \frac{\tilde{\mathcal{C}}(\mathbf{x})}{\sqrt{2}} 
\end{equation}
where $\sqrt{2}$ is the largest distance in $[0,1]^2$, and $\tilde{\mathcal{C}}(\mathbf{x})$ is the connectivity function computed on the normalized domain:
\begin{equation}
D=[0,d_x]\times[0,d_y]\implies 
\tilde{C}_j =\{(x/d_x,y/d_y):(x,y)\in C_j\}\subseteq [0,1]^2.
\end{equation}
Thus, the aggregated constraint function is
\begin{equation}
g(\mathbf{x}) = c(\tilde{g}_1(\mathbf{x}) + \tilde{g}_2(\mathbf{x}))
\end{equation}
where $c>0$ denotes a constraint penalization factor. 

To directly apply any general BBO method to this problem, we transform the constrained problem into an unconstrained one, where the objective is penalized as follows:
\begin{equation}
f_\text{obj}(\mathbf{x}) = \begin{cases}
f(\mathbf{x}) &\text{if } g(\mathbf{x}) = 0,\\
\max_{\mathbf{x}_\circ\in\mathcal{X}_\circ} f(\mathbf{x}_\circ) + g(\mathbf{x}) &\text{otherwise.}
\end{cases}
\end{equation}
Here, $\mathbf{x}_\circ\in\mathcal{X}_\circ$ is a feasible design. That is, we create an artificial funnel structure in our landscape, where the objective function value assigned to an infeasible design is lower-bounded by the one of the worst feasible design over the search space. 

To define $\mathbf{x}_\circ$, we make an informed estimate by considering several single-beam designs with horizontal orientation and varying thicknesses, from a single pixel up to the maximum allowed by the 50\% volume constraint. Among these designs, the compliance values ranged from slightly below 500 for thin beams to about 0.5 for the thickest one. Therefore, we approximate $\max_{\mathbf{x}_\circ\in\mathcal{X}_\circ} f (\mathbf{x}_\circ)\sim 500$ and set $c=10^3$, as $f$ spans roughly three orders of magnitude.

\section{Experimental Setup}

To limit the experiments' scope, we consider three design dimensionalities: 10D, 20D, and 50D. We evaluate three parameterizations: MMC, curved MMC, and HT. These represent a range of geometric flexibility: MMC constrains to straight beams (5 DoF per beam), curved MMC allows more flexible beam deformation (10 DoF per beam), and HT is a rigid fixed tiling (1 DoF per tile).

For all experiments, we consider a $(100\times50)$ design domain composed of unit cells and assume linear elastic material properties consistent with \cite{serhat2019lamination}: longitudinal and transverse Young's moduli of $E_1 = 25$ and $E_2 = 1$, respectively, an in-plane shear modulus of $G_{12} = 0.5$, and a Poisson's ratio of $\nu_{12} = 0.25$.

The geometries are first represented as polygons and are then rasterized onto the FEM mesh for simulation. Using this discretized polygonal form of the geometry, the constraints are computed using the \verb|shapely| Python package \cite{gillies_2024_13345370}.

Three optimizers are evaluated with a total simulation budget of $20D$, where $D$ denotes the dimensionality of the search space. 
This simulation budget differs from the total evaluation budget, as infeasible candidates are identified prior to simulation and can therefore be handled without consuming simulation calls. The considered optimizers are DE, CMA-ES, and HEBO.

This all results in a total of 27 experiment configurations (3 design dimensionalities $\times$ 3 parameterizations $\times$ 3 optimizers). Each configuration is run for 15 independent runs with different random seeds. For the implementation and reproducibility details, please refer to the GitHub code repository available at 
\url{https://github.com/jelle-westra/tobbo.git}.

\subsection{Parameterization Settings}
Since the Curved MMC parametrization has twice the DoF of the regular MMCs, a regular MMC with the same parameterization dimension will contain twice as many beams as the Curved MMC. For HT, each tile has a single design parameter, so the number of tiles directly corresponds to the dimensionality of the problem.

The MMC and Curved MMC parameterizations are free to move their endpoints within the full design domain. In contrast, HT is restricted to the bottom half of the design domain, which is then mirrored. We define this grid with two rows, three rows, and four rows of tiles for the 10D, 20D, and 50D designs respectively. This ensures that the tiles maintain a roughly equilateral shape.
\begin{figure}[h]
    \includegraphics[trim={0 0.7cm 0 0.7cm},clip,width=\linewidth]{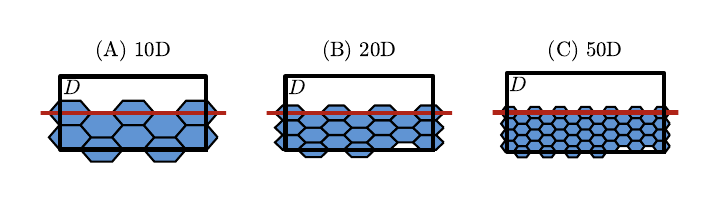}
    \caption{The HTs for dimensions: (A) 10D, (B) 20D, and (C) 50D respectively, showing the deletion of tiles: one for 20D and two for 50D.}
    \label{fig:hex-tiles}
\end{figure}
As shown in \autoref{fig:hex-tiles}, to match the dimensionalities of MMCs and Curved MMCs, in the HT parametrization we remove the least relevant tiles (the bottom row, starting from the right to preserve connectivity to the left wall). This results in deleting one tile in 20D and two in 50D.
\autoref{tab:design-dim} lists the number of total beams or tiles for each dimensionality and parameterization, after doubling the number of elements to account for the horizontal symmetry line.

\begin{table}[h]
\centering 
\caption{The total number of beams or tiles for each design dimensionality and parameterization, after mirroring about the horizontal symmetry line.}
\begin{tabular}{l|ccc}
Design Dimensionality & 10D & 20D & 50D \\
\hline
MMC Beams (after mirroring) & 4 & 8 & 20 \\
Curved MMC Beams (after mirroring) & 2 & 4 & 10 \\
HT (after mirroring) & 20 & 40 & 100 \\
\end{tabular}
\label{tab:design-dim}
\end{table}

\subsection{Optimizer Settings}
We run all optimizers with their default hyperparameter settings as provided in their respective libraries, as illustrated in \autoref{tab:optimizer-settings} alongside the respective population sizes. Additionally, as by default, for DE the (dithered) mutation factor is $(0.5,1)$, for CMA-ES the initial step-size is $\sigma^{(0)}=0.25$, and for HEBO we use the MACE acquisition function and a Matérn 5/2 kernel.

\begin{table}[h]
\centering 
\caption{Population and batch sizes for DE, CMA-ES, and HEBO per design dimensionality, and references to their respective implementations.}
\begin{tabular}{l|ccc|c}
Optimizer & 10D & 20D & 50D & \\
\hline
DE (population size) & 150 & 300 & 750 & \cite{2020SciPy-NMeth} \\
CMA-ES (population size) & 10 & 12 & 15 & \cite{nikolaus_hansen_2025_16410951} \\
HEBO (batch size) & 1 & 1 & 1 & \cite{cowen2022hebo} \\
\end{tabular}
\label{tab:optimizer-settings}
\end{table}

\section{Results}
\begin{figure}[h]
    \makebox[\textwidth][c]{\includegraphics[width=\textwidth]{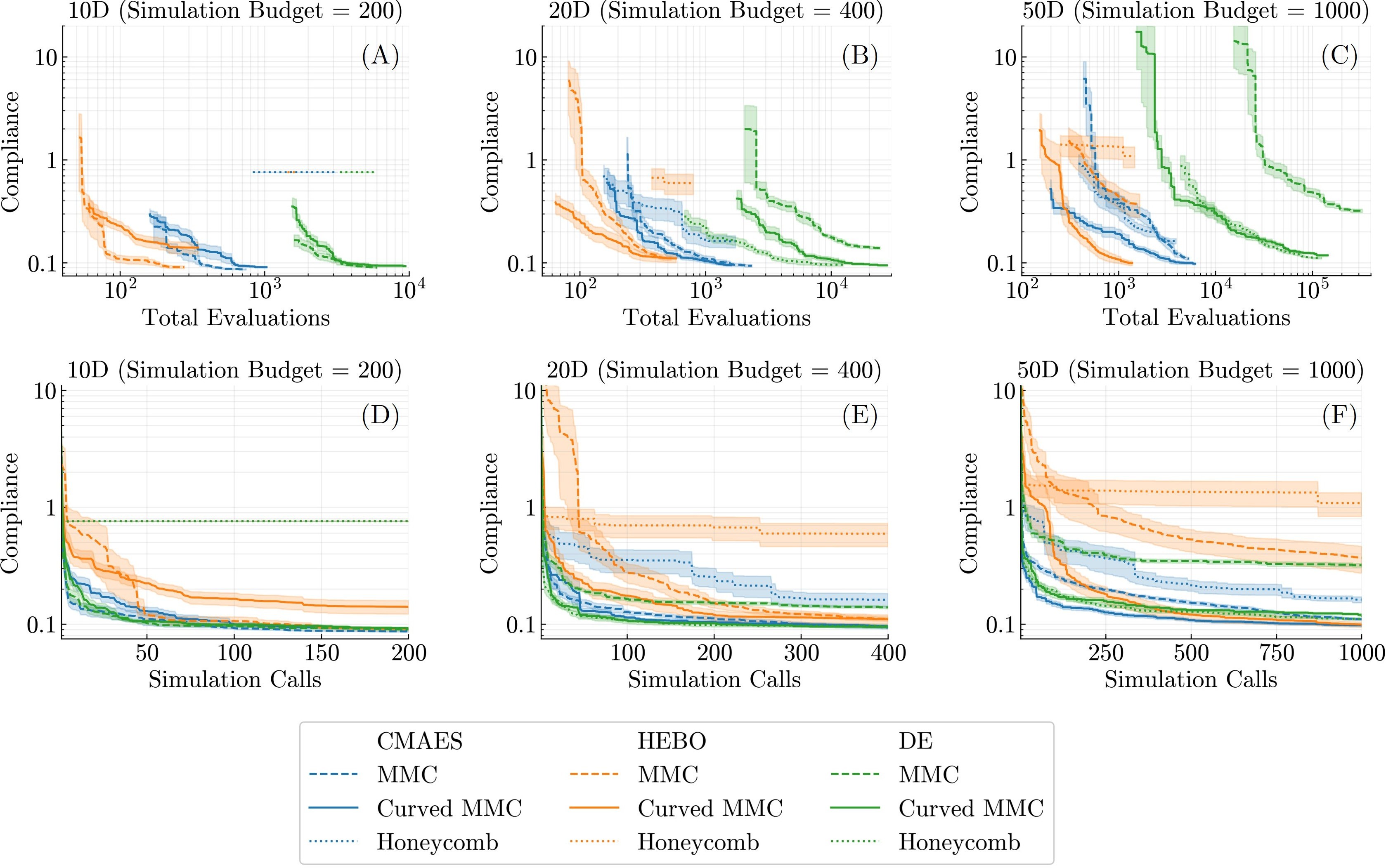}}
    \caption{Mean convergence curves for all experiment configurations (10D, 20D, and 50D), with standard error shown as shaded regions. The top row plots compliance against the total evaluation budget (including both feasible and infeasible evaluations). Here, we plot the objective value only for feasible points, as the infeasible ones are not simulated. The bottom row reports compliance with respect to the number of feasible evaluations only, i.e., simulation calls.
    }
    \label{fig:curves}
\end{figure}

As shown in Figures \ref{fig:curves}A, \ref{fig:curves}B, and \ref{fig:curves}C, we observe that, for all dimensionalities and parametrizations, HEBO uses the entire simulation budget in the smallest number of evaluations, followed by CMA-ES and then DE. HEBO is known for its sample efficiency due to its surrogate-informed sampling. 
We see that, even though CMA-ES and DE take many more evaluations to complete, they outperform HEBO within the given simulation budget (Figures \ref{fig:curves}D, \ref{fig:curves}E, and \ref{fig:curves}F).

\begin{figure}[h]
    \makebox[\textwidth][c]{\includegraphics[width=\textwidth]{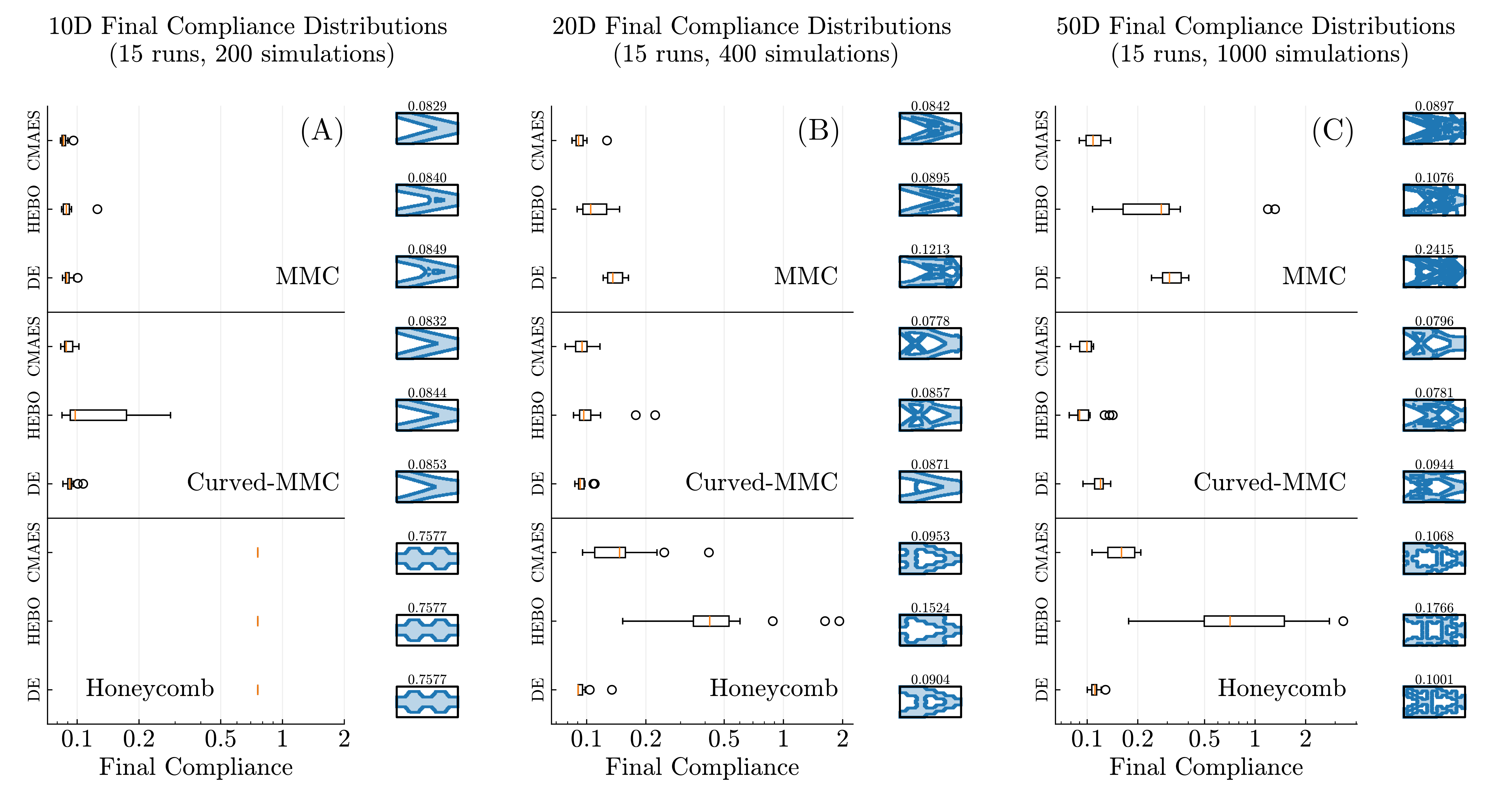}}
    \caption{Aggregated results for all final objective distributions, alongside the best found design for each configuration. Subplots (A), (B), and (C) correspond to the 10D, 20D, and 50D settings, respectively.}
    \label{fig:boxplots}
\end{figure}

In the 10D experiments (Figure~\ref{fig:curves}A and ~\ref{fig:curves}D) we observe that the regular MMC parameterization outperforms the other two parameterizations across all optimizers. This is likely due to the increased design freedom given by having double the number of beams compared to the Curved MMC parameterization. Also, as shown in Figure~\ref{fig:boxplots}A, all algorithms converge to a V-shape design, which is a local optimum and is particularly suited by the straight beams parameterization. We see similar V-shaped designs with the curved MMCs, however
HEBO shows higher variability in the results and a distribution shifted to the right compared to the other algorithms.
The Curved MMCs have more complex interaction among the design variables (10 DoF per beam) in comparison to the regular MMC parameterization (5 DoF per beam). It could be the case that too many simulation calls are wasted modeling these interactions in the surrogate model for Curved MMCs. On average, it takes HEBO only 50 infeasible calls to find feasible designs with the MMC parameterizations. This is the case even though, based on our preliminary investigations, the feasible region accounts only for about 0.005\% of the total design space for the 10D Curved MMC parametrization, details of which are available in the code repository. 
In 10D, the HT parameterization performs the worst: the design is so restricted that all the optimizers converge to exactly the same, suboptimal, design. By observing all $2^{10}$ generated configurations, we identified this as the only feasible design.

For 20D (Figure~\ref{fig:curves}B and \ref{fig:curves}E), dominance gradually shifts from the MMC to the Curved MMC parametrization. At the same time, optimizer-dependent differences become more pronounced for HT and emerge for MMC, while Curved MMC remains comparatively robust across different optimizers, despite some outliers for HEBO (Figure~\ref{fig:boxplots}B). Given that regular MMCs have double the number of beams, the complexity of the design space starts to play a role here. Again, we note the following: the best parameterization gives competitive performance across all our selected optimizers.

The HT parameterization was deliberately chosen as a challenging case for continuous optimizers. Due to its static cell structure and hard thresholding, it can represent far less diverse and complex designs compared to the MMC parameterizations. With such a restrictive representation, the choice of optimizer has the strongest impact on final performance, confirming that optimizer effectiveness depends on the quality of the parameterization. Notably, DE shows exceptional performance under HT, achieving solution qualities close to the best results obtained with the other parameterizations. In this simplistic setting, performance differences among algorithms are the most pronounced, suggesting that some optimizers are better suited to navigating the discontinuous, piecewise-constant landscape induced by the HT parameterization. 

For 50D (Figure~\ref{fig:curves}C and~\ref{fig:curves}F, Figure~\ref{fig:boxplots}C) the trend continues, the added complexity from having a total of 20 beams in regular MMC in comparison to only 10 beams in Curved MMCs does not benefit the optimizers. For both MMC and HT, the ranking of the optimizers stays the same as in the 20D case, but the performance gap increases.
The observed performance degradation of HEBO can be explained by the combinatorial explosion of symmetries in the design space. Beam permutations and endpoint swaps induce $m! \cdot 2^{2m}$ equivalent configurations for a system of $m$ beams, generating a highly multi-modal objective landscape, which is challenging to model for the surrogate.
This symmetry group grows factorially: for $m=5$ beams this results in 122,880 symmetric states and for $m=10$ this increases to 3,805,072,588,800. These correspond to the Curved MMC and standard MMC cases, respectively, in our 50D experiments. Global BO methods like HEBO are disproportionately affected in such landscapes, whereas CMA-ES and DE can exploit localized structure more efficiently.
Hence, we note the overall pattern: a state-of-the-art optimizer does not fix a poor parameterization of the problem, but a well-chosen parameterization reduces reliance on algorithm selection for competitive performance. 

To statistically validate these observations, we performed pairwise Mann-Whitney U tests between all optimizers for each parameterization and dimensionality combination. The resulting p-values are presented in \autoref{tab:mwu}. 
Although we note similarity in performance among many optimizers in the 10D MMC case, we still find statistical evidence for difference among distributions of the CMA-ES and  DE optimizers.  The difficulty HEBO has with the Curved MMC is emphasized here in comparison to CMA-ES. In 20D, all p-values exceed 0.05 for the Curved MMC parametrization, indicating no significant differences among optimizers. This suggests that this parameterization is particularly well-suited to the problem, leading to reduced sensitivity to the choice of optimization algorithm. 
For 50D, we observe the pattern noted earlier: the need for careful algorithm selection becomes more pronounced. Although Curved MMC remains the most robust algorithm choice, a ranking begins to emerge, with only CMA-ES and HEBO performing equally well.


\begin{table}[H]
\centering
\caption{P-values from Mann–Whitney U tests comparing optimizer performance across parameterizations and dimensionalities. Bold values indicate cases where the null hypothesis of equal performance cannot be rejected ($p \geq 0.05$).}
\label{tab:mwu}
\resizebox{\textwidth}{!}{%
\begin{tabular}{rcl|ccc|ccc|ccc}
\toprule
& & & &10D & &&20D & &&50D & \\
\midrule
& &  &MMC&Curved&Honey-& MMC&Curved&Honey-& MMC&Curved&Honey- \\
& &  & &MMC&comb& &MMC&comb& &MMC&comb \\
\midrule
CMA-ES&-&HEBO& \textbf{0.0620} & 0.0202 & \textbf{1.0000} & 0.0021  & \textbf{0.3615} & <0.0001 & <0.0001 & \textbf{0.3401} & <0.0001 \\
CMA-ES&-&DE  & 0.0101 & \textbf{0.2998} & \textbf{1.0000} & <0.0001 & \textbf{1.0000} & <0.0001 & <0.0001 & 0.0002 & <0.0001 \\
HEBO&-&DE    & \textbf{0.6783} & \textbf{0.0512} & \textbf{1.0000} & 0.0005  & \textbf{0.2628}& <0.0001 &  \textbf{0.0680} & 0.0048 & <0.0001 \\
\bottomrule
\end{tabular}
}
\end{table}

\section{Conclusions and Future Work}

This paper investigated the critical interplay between design parameterization and black-box optimization algorithm performance in constrained structural topology optimization. Our case study involved minimizing compliance for a horizontal cantilever, using three distinct parameteri\-zations---MMCs, Curved MMCs, and Honeycomb Tiling---and three black-box optimization
optimizers---CMA-ES, DE, and HEBO---across 10D, 20D, and 50D problems.

Our primary finding is that the quality of the parameterization exerted a more significant impact on the final solution quality than the choice of black-box optimization optimizer across all tested dimensions.
For high-quality parameterizations (e.g., MMCs in 10D, and Curved MMCs in 20D/50D), all three tested optimizers yielded competitive performance. This suggests that a well-structured design space can significantly enhance the robustness of the optimization process, making it less sensitive to the specific search strategy employed or its fine-tuning.
Conversely, poorly structured parameterizations (e.g., honeycomb tiling or high number of MMC beams) led to substantial performance disparities among the optimizers, indicating a heavy reliance on the specific optimizer choice and tuning for competitive results.

Despite the cantilever beam case study considered in our analysis is a well-known problem solvable by gradient-based methods, the parameterizations and optimizers we compared are not limited to this simple scenario. Therefore, the dominance of parameterization quality over optimizer choice is likely relevant to a broader class of constrained, high-dimensional black-box optimization problems in engineering design.
Hence, for highly constrained real-world topology optimization problems, the effort invested in thoughtful, task-specific parameterization design may still yield a greater return on performance than extensive optimizer selection or hyperparameter tuning. 

To confirm the generality of our findings, we plan to extend this study to a broader range of optimizers, including additional evolution strategies and specialized Bayesian optimization variants, as well as alternative parameterizations. It would also be valuable to investigate whether the observed trends persist across different structural optimization problems and in entirely distinct application domains, such as crashworthiness design, which inspired our analysis.



\begin{credits}

\subsubsection{\discintname}
The authors have no competing interests to declare that are
relevant to the content of this article.

\end{credits}
%
%
%
\bibliographystyle{splncs04}
\bibliography{main}

\begin{thebibliography}{10}
\providecommand{\url}[1]{\texttt{#1}}
\providecommand{\urlprefix}{URL }
\providecommand{\doi}[1]{https://doi.org/#1}

\bibitem{andreassen2011efficient}
Andreassen, E., Clausen, A., Schevenels, M., Lazarov, B.S., Sigmund, O.: Efficient topology optimization in matlab using 88 lines of code. Structural and Multidisciplinary Optimization  \textbf{43}(1),  1--16 (2011)

\bibitem{back1996evolutionary}
B\"{a}ck, T.: Evolutionary algorithms in theory and practice: evolution strategies, evolutionary programming, genetic algorithms. Oxford university press (1996)

\bibitem{back1997handbook}
B{\"a}ck, T., Fogel, D.B., Michalewicz, Z.: Handbook of evolutionary computation. Release  \textbf{97}(1), ~B1 (1997)

\bibitem{bendsoe1989optimal}
Bends{\o}e, M.P.: Optimal shape design as a material distribution problem. Structural optimization  \textbf{1},  193--202 (1989)

\bibitem{bendsoe2013topology}
Bendsoe, M.P., Sigmund, O.: Topology optimization: theory, methods, and applications. Springer Science \& Business Media (2013)

\bibitem{bujny2020level}
Bujny, M.: Level set topology optimization for crashworthiness using evolutionary algorithms and machine learning. Ph.D. thesis, Technische Universit{\"a}t M{\"u}nchen (2020)

\bibitem{bujny2018identification}
Bujny, M., Aulig, N., Olhofer, M., Duddeck, F.: Identification of optimal topologies for crashworthiness with the evolutionary level set method. International Journal of Crashworthiness  \textbf{23}(4),  395--416 (2018)

\bibitem{chapman1994genetic}
Chapman, C.D., Saitou, K., Jakiela, M.J.: Genetic algorithms as an approach to configuration and topology design. Journal of mechanical design  \textbf{116}(4),  1005--1012 (1994)

\bibitem{cowen2022hebo}
Cowen-Rivers, A.I., Lyu, W., Tutunov, R., Wang, Z., Grosnit, A., Griffiths, R.R., Maraval, A.M., Jianye, H., Wang, J., Peters, J., et~al.: Hebo: Pushing the limits of sample-efficient hyper-parameter optimisation. Journal of Artificial Intelligence Research  \textbf{74},  1269--1349 (2022)

\bibitem{forresterEngineeringDesignSurrogate2008}
Forrester, A.I.J., S{\'o}bester, A., Keane, A.J.: Engineering {{Design}} via {{Surrogate Modelling}} - {{A Practical Guide}}. John Wiley \& Sons Ltd. (2008)

\bibitem{garnett2023bayesian}
Garnett, R.: Bayesian optimization. Cambridge University Press (2023)

\bibitem{gillies_2024_13345370}
Gillies, S., van~der Wel, C., Van~den Bossche, J., Taves, M.W., Arnott, J., Ward, B.C., et~al.: Shapely (2024)

\bibitem{guo2016explicit}
Guo, X., Zhang, W., Zhang, J., Yuan, J.: Explicit structural topology optimization based on moving morphable components (mmc) with curved skeletons. Computer methods in applied mechanics and engineering  \textbf{310},  711--748 (2016)

\bibitem{guo2014doing}
Guo, X., Zhang, W., Zhong, W.: Doing topology optimization explicitly and geometrically—a new moving morphable components based framework. Journal of Applied Mechanics  \textbf{81}(8),  081009 (2014)

\bibitem{hansen2009real}
Hansen, N., Finck, S., Ros, R., Auger, A.: Real-parameter black-box optimization benchmarking 2009: Noiseless functions definitions. Ph.D. thesis, INRIA (2009)

\bibitem{hansen2001completely}
Hansen, N., Ostermeier, A.: Completely derandomized self-adaptation in evolution strategies. Evolutionary computation  \textbf{9}(2),  159--195 (2001)

\bibitem{nikolaus_hansen_2025_16410951}
Hansen, N., Ueno, Y., ARF1, Cakmak, S., Kadlecová, G., Abad~López, G., Nozawa, K., Rolshoven, L., Akimoto, Y., brieglhostis, Brockhoff, D.: Cma-es/pycma: v4.3.0 (2025)

\bibitem{hutterAutomatedMachineLearning2019a}
Hutter, F., Kotthoff, L., Vanschoren, J.: Automated Machine Learning: Methods, Systems, Challenges. Springer Nature (2019)

\bibitem{jinSurrogateassistedEvolutionaryComputation2011}
Jin, Y.: Surrogate-assisted evolutionary computation: {{Recent}} advances and future challenges. Swarm and Evolutionary Computation  \textbf{1}(2),  61--70 (2011)

\bibitem{kerschkeAutomatedAlgorithmSelection2019a}
Kerschke, P., Hoos, H.H., Neumann, F., Trautmann, H.: Automated {{Algorithm Selection}}: {{Survey}} and {{Perspectives}}. Evolutionary Computation  \textbf{27}(1),  3--45 (2019)

\bibitem{kruskal1956shortest}
Kruskal, J.B.: On the shortest spanning subtree of a graph and the traveling salesman problem. Proceedings of the American Mathematical society  \textbf{7}(1),  48--50 (1956)

\bibitem{longGeneratingCheapRepresentative2024}
Long, F.X., Van~Stein, N., Frenzel, M., Krause, P., Gitterle, M., B{\"a}ck, T.: Generating {{Cheap Representative Functions}} for {{Expensive Automotive Crashworthiness Optimization}}. ACM Transactions on Evolutionary Learning and Optimization  \textbf{4}(2),  1--26 (2024)

\bibitem{mdobook}
Martins, J.R.R.A., Ning, A.: Engineering Design Optimization. Cambridge University Press, Cambridge, UK (2022), \url{https://mdobook.github.io}

\bibitem{mersmann2011exploratory}
Mersmann, O., Bischl, B., Trautmann, H., Preuss, M., Weihs, C., Rudolph, G.: Exploratory landscape analysis. In: Proceedings of the 13th annual conference on Genetic and evolutionary computation. pp. 829--836 (2011)

\bibitem{raponi2019kriging}
Raponi, E., Bujny, M., Olhofer, M., Aulig, N., Boria, S., Duddeck, F.: Kriging-assisted topology optimization of crash structures. Computer Methods in Applied Mechanics and Engineering  \textbf{348},  730--752 (2019)

\bibitem{saxena2003honeycomb}
Saxena, R., Saxena, A.: On honeycomb parameterization for topology optimization of compliant mechanisms. In: International Design Engineering Technical Conferences and Computers and Information in Engineering Conference. vol. 37009, pp. 975--985 (2003)

\bibitem{serhat2019lamination}
Serhat, G., Basdogan, I.: Lamination parameter interpolation method for design of manufacturable variable-stiffness composite panels. AIAA Journal  \textbf{57}(7),  3052--3065 (2019)

\bibitem{storn1997differential}
Storn, R., Price, K.: Differential evolution--a simple and efficient heuristic for global optimization over continuous spaces. Journal of global optimization  \textbf{11}(4),  341--359 (1997)

\bibitem{svanberg1987method}
Svanberg, K.: The method of moving asymptotes—a new method for structural optimization. International journal for numerical methods in engineering  \textbf{24}(2),  359--373 (1987)

\bibitem{2020SciPy-NMeth}
Virtanen, P., Gommers, R., Oliphant, T.E., Haberland, M., Reddy, T., Cournapeau, D., Burovski, E., Peterson, P., Weckesser, W., Bright, J., {van der Walt}, S.J., Brett, M., Wilson, J., Millman, K.J., Mayorov, N., Nelson, A.R.J., Jones, E., Kern, R., Larson, E., Carey, C.J., Polat, {\.I}., Feng, Y., Moore, E.W., {VanderPlas}, J., Laxalde, D., Perktold, J., Cimrman, R., Henriksen, I., Quintero, E.A., Harris, C.R., Archibald, A.M., Ribeiro, A.H., Pedregosa, F., {van Mulbregt}, P., {SciPy 1.0 Contributors}: {{SciPy} 1.0: Fundamental Algorithms for Scientific Computing in Python}. Nature Methods  \textbf{17},  261--272 (2020)

\end{thebibliography}





\end{document}